\pdfoutput=1

\documentclass[11pt]{article}

\usepackage{emnlp2021}

\usepackage{times}
\usepackage{latexsym}
\usepackage{comment}
\usepackage{tikz}

\usepackage[T1]{fontenc}

\usepackage[utf8]{inputenc}

\usepackage{microtype}

\title{A Simple Post-Processing Technique for Improving Readability Assessment of Texts using Word Mover's Distance}

\author{Joseph Marvin Imperial\textsuperscript{1} \and Ethel Ong\textsuperscript{2} \\
  \textsuperscript{1}National University, \textsuperscript{2}De La Salle University \\
  Manila, Philippines \\
  \texttt{jrimperial@national-u.edu.ph} \\
  }

\begin{document}
\maketitle
\begin{abstract}
Assessing the difficulty level of reading materials or texts in general is the first step towards effective comprehension and learning. In this study, we improve the conventional methodology of automatic readability assessment by incorporating the Word Mover's Distance (WMD) of ranked texts as an additional post-processing technique to \textit{correct} the readability level output of a given trained classification model. Results of our experiments show that the proposed post-processing technique is language-agnostic, works in a multiclass setting, and outperforms previous approaches for readability assessment using three datasets in English, German, and Filipino\footnote{We will release the code upon publication.}.

\end{abstract}

\section{Introduction}
Reading is one of the important life skills that a learner has to continuously practice and improve beginning grade school. When a prescribed reading material exceeds the reading capability of a learner, they may become frustrated, disinterested, and lose confidence in their reading abilities \cite{Cambria2010,Hasyim2018}. Readability assessment is the process used by linguists and educators to evaluate the ease or difficulty of texts with the intent of prescribing reading materials to learners that are appropriate to their abilities. 

Automated readability assessment can be viewed as a supervised task where annotated data is required and thus can be approached generally in two ways: regression and classification. In regression-based approaches \cite{flor-etal-2013-lexical,Guevarra2011}, the readability level of books is provided as a value from a fixed range. This provides more fine-grained information about the magnitude of readability of texts. However, recent works have favored classification-based models \cite{chatzipanagiotidis-etal-2021-broad,weiss-meurers-2018-modeling,xia-etal-2016-text,reynolds-2016-insights,hancke-etal-2012-readability,vajjala-meurers-2012-improving} for their simplicity and ease in collecting gold-standard data as experts usually give annotations by category of difficulty \cite{deutsch-etal-2020-linguistic}.

In this paper, we propose the use of Word Mover's Distance (WMD) as an additional post-processing task for automatic readability assessment. WMD, a novel distance function developed by \citet{pmlr-v37-kusnerb15} using word embeddings, allows identification of dissimilarity between two texts by calculating the distance or the amount of effort to transport all words from one document to another within the word embedding space. Thus, we argue that the shorter the distance between two texts, the similar they are in terms of readability and vice versa. Likewise, the application of a post-processing technique acts as a \textit{second-tier reduction process} by minimizing the errors made of a trained classification model by considering the distribution of words among similar texts in the same category of reading level. We test the proposed methodology on top of a trained SVM model using three monolingual datasets in Filipino, English, and German. We also provide an empirical evaluation of the model's performance and limitations with other approaches in readability assessment.

\section{Preliminaries}

The Word Mover's Distance (WMD) is a word embedding-based distance function by \citet{pmlr-v37-kusnerb15} inspired by the Earth Mover's Distance \cite{rubner1998metric} transportation problem. WMD calculates the dissimilarity between texts in terms of Euclidean distance in the word embedding space. The \textit{travel cost} of texts (represented as bag-of-words vectors) is calculated as the minimized cumulative aggregation cost of \textit{moving} words from a source text $D_i$ to a destination text $D_j$. The model can be formally viewed as follows,

\begin{equation}
\min_{\mathbf{T}\geq 0} \sum_{i,j=1}^{n} = \mathbf{T}_{i,j} c(i,j)
\end{equation}

where $n$ is the total vocabulary size, $c(i,j) = \left \| x_i - x_j \right \|$ is the distance between two words in the semantic space, and $\mathbf{T}\geq 0$ denotes the converted sparse flow matrix of how much words from the source text $D_i$ travels to the destination text $D_j$. 


WMD is a semantic-based approach for estimating the (dis)similarity of documents. However, we hypothesize that WMD can also capture the readability of documents  as long as the documents being compared do not have the same, exact meaning. We leverage our hypothesis on the common knowledge that documents with a high level of reading difficulty will often use complex words and documents with a low level reading difficulty will often use simple, easy words regardless of genre. For example, difficult or challenging words that may be encountered in college textbooks are considered high level and most likely will not be present in low level reading materials such as children's story and picture books. Thus, if documents $D_1$ and $D_2$, where $D_2$ is a document of higher reading level than $D_1$, then the WMD distance of a target document $D_0$ with unknown reading level is shorter with respect to $D_2$ if they belong to the same category of reading difficulty and share similar words.

To ground this hypothesis, we use the OneStopEnglish corpus \cite{vajjala-lucic-2018-onestopenglish} which is composed of open-source English reading materials\footnote{Additional description found in Section 4.}. The purpose of this test is to identify if the WMD scores of documents with distinct readability levels do not coincide with documents with distinct semantics and is statistically significant in terms of difference. From the corpus, we handpicked and aggregated two groups: (a) the first group being 189 pairs of documents with distinct reading levels and (b) the second group being 189 pairs of documents belonging to the same reading level. Documents from both groups are semantically different. We obtained the WMD scores of each pair for each group and performed a two-sample Mann-Whitney U test. Results showed that the difference between WMD scores between readability ($\mu$ = 0.089) and semantics ($\mu$ = 0.041) of documents is statistically significant ($p$ < 0.05)\footnote{test $Z$ statistic = 16.214 not in the 95\% critical value accepted range: [-1.960,1.960]}, proving our initial hypothesis.

\section{Post-Processing Technique with WMD}
\label{postProcessing}
The variation of this approach from previous works \cite{ma-etal-2012-ranking} is in the method of ranking texts and the procedure used for transforming relative order to a discrete value. Our proposed post-processing technique involves three major phases: classification, ranking, and grounding.

For the \textbf{classification phase}, a model is trained using vanilla SVM (based from the commonly-used models mentioned in previous works) using hand-crafted linguistic feature sets to produce a series of probabilities for each instance of the training data per corresponding class. The target text where the readability level is needed will also be subjected to the model. Each text in the training data as well as the target text will now have a collection of probabilities for each class. 

In the \textbf{ranking phase}, texts from the training data plus the target text will then be ranked in an increasing order with respect to the probability of the most difficult class in terms of readability to produce a collection of sorted texts or a \textit{bookshelf} arranged from easiest to most difficult. Each instance of the training data in this phase acts as a \textit{guide book} wherein it will preemptively help identify the readability level of the target text since it is assumed that the bookshelf is arranged according to readability.

\begin{figure}[!htbp]
    \centering
    \includegraphics[width=0.45\textwidth]{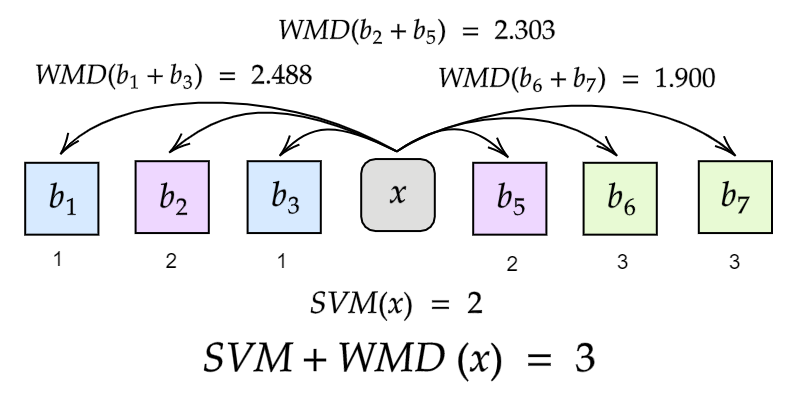}
    \caption{Label evaluation in the case of triple tie. The class with the least, normalized distance from the target text $x$ will be selected, in this case it is 3.}
    \label{fig:tie}
\end{figure}

In the \textbf{grounding phase}, the location of the target text is identified from the sorted bookshelf of texts. The distribution of words of the target text will be compared with its first three neighboring texts from left and right with WMD. To get the final readability level, a hard-voting approach will be performed to get the majority readability level from the three compared neighbors bidirectionally. In case of a tie, the neighbor with the least, normalized WMD difference (since the algorithm is distance-based) to the target text will be selected for the corrected label. Figure~\ref{fig:tie} visually describes how the post-processing technique operates in the case of a three-way tie.


\section{Data Description}
We test the proposed post-processing method on three datasets composed of leveled texts in English, German, and Filipino as described in Table~\ref{tab:data}. For implementation procedures, we obtained the code from the authors for the extraction of the linguistic feature sets spanning traditional, lexical, syntactic, psycholinguistic, and morphological used to train readability assessment models. We extracted 165, 155, and 54 features as done in previous works for German \cite{hancke-etal-2012-readability}, English \cite{vajjala-lucic-2018-onestopenglish}, and Filipino \cite{Imperial2019,imperial2020exploring,imperial2021application} respectively. \linebreak

\noindent\textbf{GEO--GEOlino corpus.} The GEO--GEOlino corpus is an open-source dataset containing 4,599 magazine articles in German from the work of \citet{hancke-etal-2012-readability} and \citet{weiss-meurers-2018-modeling}. These are frequently used as baseline for German-related readability assessment. The articles in the GEO corpus cover the domains of nature, culture, and science published by Gruner and Jahr\footnote{https://www.geo.de} while GEOlino contains topics on handicraft, games, and animals targeted for children age 8-14 years old. \linebreak

\noindent\textbf{OneStopEnglish (OSE) corpus.} The OSE corpus is a collection of English texts in three levels of difficulty (beginner, intermediate, and advanced) for adult ESL learners. The corpus was compiled by \citet{vajjala-lucic-2018-onestopenglish} from the OneStopEnglish learning resource website\footnote{https://www.onestopenglish.com}. A total of 567 texts were used for this study.
\linebreak

\noindent\textbf{Adarna House corpus.} The Adarna House data is a collection of Filipino story books published by Adarna House Inc.\footnote{https://adarna.com.ph/} that are commonly used in the general educational sector of the Philippines. A total of 265 reading materials distributed into three levels (beginner, intermediate, and advanced) were used for the study.
\linebreak


\begin{table}[!htbp]\small
\centering
\begin{tabular}{|l|c|c|c|}

\hline \bf Data & \bf Doc Count & \bf \# Class &\bf  Vocab\\  \hline

GEO-GEOlino & 4,599 & 2 & 191,262\\ \hline
OneStopEnglish & 567 & 3 & 17,818\\ \hline
Adarna House & 265 & 3 & 16,058\\ \hline

\end{tabular}
\caption{\label{tab:data}
Data distribution for English, German, and Filipino.}
\end{table}

\section{Experiment Setup}
We trained an SVM model using the various feature sets extracted for each corresponding dataset used. We chose SVM due to its significant usage in previous works \cite{chatzipanagiotidis-etal-2021-broad,deutsch-etal-2020-linguistic,imperial2020exploring,weiss-meurers-2018-modeling,hancke-etal-2012-readability}. We used a 5-fold cross validation procedure for training the models. After training, we applied the proposed post-processing technique described in Section \ref{postProcessing} and evaluated the performance using accuracy and F1 with the test split. For the word embeddings of English, German, and Filipino needed for the technique, we downloaded the resources from the fastText website\footnote{https://fasttext.cc/docs/en/crawl-vectors.html}. The word embeddings in various languages were trained from Common Crawl and Wikipedia datasets by \citet{grave-etal-2019-training}. 



\begin{table*}[!htbp]\small
\centering
\begin{tabular}{|l|c|c|c|c|c|c|}
\hline

\multicolumn{1}{|c|}{\textbf{Method}} & 
\multicolumn{2}{c|}{\textbf{GEO-GEOlino}} & 
\multicolumn{2}{c|}{\textbf{OneStopEnglish}} & 
\multicolumn{2}{c|}{\textbf{Adarna House}} \\ \cline{2-7} 

\multicolumn{1}{|c|}{} & 
\multicolumn{1}{c|}{\textbf{Acc}} & \multicolumn{1}{c|}{\textbf{F1}} & \multicolumn{1}{c|}{\textbf{Acc}} & \multicolumn{1}{c|}{\textbf{F1}} & \multicolumn{1}{c|}{\textbf{Acc}} & \multicolumn{1}{c|}{\textbf{F1}} \\ \hline

Vanilla SVM & 0.813 & 0.799 & 0.711 & 0.694 & 0.413 & 0.416 \\ \hline

Binary Insertion Sort Ranking & {0.812} &  {0.797} &  {0.659} & {0.667} & {0.444} & {0.357} \\ 
\cite{tanaka-ishii-etal-2010-sorting} & & & & & & \\ \hline

RankSVM w/ 3-Neighbor Scheme & {0.867} &  {0.861} &  {0.658} & {0.644} & {0.365} & {0.359} \\ 
\cite{ma-etal-2012-ranking} & & & & & & \\ \hline

\bf SVM w/ Proposed WMD Technique & \bf 0.890 & \bf 0.892 & \bf 0.746 & \bf 0.749 & \bf 0.538 & \bf 0.538  \\ \hline

\end{tabular}
\caption{\label{experiments}
The accuracy and F1 scores of the proposed post-processing technique with WMD on top of a trained SVM model outperformed previous ranking-based approaches in three monolingual datasets.}
\end{table*}

\section{Results and Discussion}
We compared the performance of the proposed technique against three previous approaches: (a) a trained vanilla SVM model, (b) Binary Insertion Sort ranking by \citet{tanaka-ishii-etal-2010-sorting}, the first ranking method applied in readability assessment, and (c) a RankSVM model by \citet{ma-etal-2012-ranking} which also uses a bidirectional three-neighbor scheme. We especially highlight the work of \citet{ma-etal-2012-ranking} as it is the one closest to our approach for analysis; the difference in our approach is that we consider the actual text contents of the documents through WMD instead of just looking at the discrete labels. 

Table~\ref{experiments} shows results of the proposed technique improving the performances of conventional classification-based model such as SVM in terms of accuracy and F1 scores in all three languages. The accuracy score provides an overview of each model's performance while the F1 values  show the harmonic mean between the models' precision and recall. As can be seen from the table, the most notable result is the largest increase when the post-processing technique is applied to the Filipino dataset with $\approx$13\% in accuracy and F1. For English and German, though not as large as the performance increase in Filipino, the post-processing technique still managed to \textit{correct} a few misclassified instances with around 2.9\% and 7.7\% for accuracy and 5.5\%  and 9.3\% for F1. We attribute the improvement in performance specifically to the ranking phase wherein each instance of the training data is used as a \textit{guide book} to place the target text in its correct position in the \textit{bookshelf} through sorting. From this, since the comparator for the ranking phase utilized the output of the trained SVM model, it is less likely to make mistakes because the SVM model itself was developed using the training data. 

With the Filipino dataset having the largest increase in performance, this means that the post-processing technique made a substantial number of corrections of labels during the ranking and grounding phases of the technique and despite the subpar performance of the vanilla SVM model. To quantify this value, the occurrence of the post-processing technique invoking the WMD tie-breaker method of normalizing distances for each readability class are 13.3\%, 10\%, and 8.7\% for Filipino, German, and English respectively. These values observe a direct relationship with the improvements of upgrading the SVM model with the WMD-based post-processing technique.  

\section{Related Work}
\label{RRL}
In automatic readability assessment, traditional machine learning models are often used rather than neural network-based models for languages except English due to limited resource of annotated data. Recent works considered modelling the hierarchical and sequential nature of documents \cite{meng2020readnet,deutsch-etal-2020-linguistic,martinc2019supervised,azpiazu-pera-2019-multiattentive}. In \citet{deutsch-etal-2020-linguistic}, the performance of models trained using SVM and Logistic Regression remains closely at par with CNN, hierarchical attention models, and Transformer-based models. Thus, using these traditional models are still practical for the field. 

Our proposed post-processing technique was inspired by ranking-based approaches which are often done for low-resource languages. \citet{tanaka-ishii-etal-2010-sorting} made a workaround with the inadequate data problem in Japanese texts by using an SVM model as a comparator to determine the relative ease of texts rather than based on an indicator of the absolute difficulty. \citet{dellorletta-etal-2012-genre} used the combined cosine distance of a target text from two poles (feature vectors extracted from easy-to-read and difficult-to-read corpus) and then performed ranking based on the accumulated distance value. \citet{vajjala-meurers-2014-assessing} used a ranking-based approach on identifying the relative difficulty of sentence pairs obtained from Wikipedia and Simple Wikipedia websites.

\section{Conclusion}
Readability assessment is the process of automating how language experts evaluate the ease or difficulty of texts. In this study, we devised a post-processing algorithm using Word Mover's Distance which can be added to classification-based readability assessment models to improve their performances for the task by correcting labels through a majority-voting scheme. The approach is (a) language-agnostic as tested with the English, German, and Filipino datasets and (b) works in a multiclass setting. Results from experiments show that the addition of the post-processing technique significantly improves performances of SVM especially for low-resource languages such as Filipino where neural-based approaches are not feasible. Further directions of the study include using knowledge-enriched word embeddings as well as more explorations in other low-resource languages.

\section{Impact Statement}
We discuss the broad impact statement and ethical considerations of the study. First, the study does not introduce a new dataset nor does it involve any human subjects. It covers only a small, focused and interesting application of a new post-processing technique in improving readability assessment models of texts in various languages. In general, the intended use of readability assessment models and its related studies are to help linguists, educators, and publishers in doing away with manual and formula-based approaches that are often shallow and time-consuming. We emphasize that it is the language experts in the field who will have a final verdict on the proper readability level of texts especially in the education sector.

\bibliography{anthology,custom}
\bibliographystyle{acl_natbib}

\end{document}